# Assessing Economic Viability: A Comparative Analysis of Total Cost of Ownership for Domain-Adapted Large Language Models versus State-of-the-art Counterparts in Chip Design Coding Assistance

Amit Sharma, Teodor-Dumitru Ene, Kishor Kunal, Mingjie Liu, Zafar Hasan and Haoxing Ren
NVIDIA Corporation
amitsharma@nvidia.com

*Abstract*—This paper presents a comparative analysis of total cost of ownership (TCO) and performance between domain-adapted large language models (LLM) and state-of-the-art (SoTA) LLMs , with a particular emphasis on tasks related to coding assistance for chip design. We examine the TCO and performance metrics of a domain-adaptive LLM, ChipNeMo, against two leading LLMs, Claude 3 Opus and ChatGPT-4 Turbo, to assess their efficacy in chip design coding generation. Through a detailed evaluation of the accuracy of the model, training methodologies, and operational expenditures, this study aims to provide stakeholders with critical information to select the most economically viable and performance-efficient solutions for their specific needs. Our results underscore the benefits of employing domain-adapted models, such as ChipNeMo, that demonstrate improved performance at significantly reduced costs compared to their general-purpose counterparts. In particular, we reveal the potential of domain-adapted LLMs to decrease TCO by approximately 90% - 95%, with the cost advantages becoming increasingly evident as the deployment scale expands. With expansion of deployment, the cost benefits of ChipNeMo become more pronounced, making domain-adaptive LLMs an attractive option for organizations with substantial coding needs supported by LLMs.

*Index Terms*—Hardware Design, CAD, LLM, TCO, Chip Design

## I. INTRODUCTION

Rapid progression in chip design technology has ushered in a new era of complexity that requires the integration of innovative solutions to refine the design workflow. In this context, automated coding and scripting mechanisms for Electronic Design Automation (EDA) applications have become indispensable to increase the efficiency of chip designers and ensure stringent code quality benchmarks [1], [2].LLM, renowned for their proficiency in mimicking human-like text, have shown significant promise in facilitating coding operations, including code completion, bug identification, and document generation [3], [4]. However, the integration of LLMs into chip design coding support presents its own set of challenges [5],[6].Furthermore, the efficacy of LLM in specialized tasks, such as chip design coding, depends on the architecture of the model, the quality and relevance of the training data, and the precision of finetuning methodologies [7].

Recent studies by Thakur et al. [8] demonstrated that open-source LLMs (such as CodeGen [9]), when fine-tuned with additional Verilog data, can surpass the performance of the latest OpenAI GPT-3.5 models in the hardware design domain.Liu et al.(2023) [10] introduce VerilogEval, a benchmarking framework designed to assess the performance of LLMs in generating Verilog code for hardware design and verification tasks. ChatEDA [11] showcased the application of LLMs in generating scripts for EDA tools, revealing that a fine-tuned LLaMA2 70B model surpasses the performance of the GPT-4 model in this specific task.

Liu et al. (2024) [12] showcase the benefits of DAPT and retrieval-augmented generation (RAG) for LLMs in the generation of EDA scripts, leading to improved performance in domain-specific tasks compared to base-based LLMs, without compromising generic capabilities. However, their work lacks a comprehensive analysis of Total Cost of Ownership (TCO) and performance compared to state-of-the-art (SOTA) larger LLMs for EDA tools and various usage workloads. Our research paper addresses this gap by providing a novel perspective on selecting appropriate LLMs based on specific software usage workload and computational cost considerations.

In this research paper, we make the following key contributions and findings regarding the selection of Large Language Models (LLMs) for software coding assistance tasks:

- We conduct a comprehensive comparative analysis of Domain Adaptive Pre-training (DAPT) models and SOTA LLMs with large parameters, focusing on their performance and computational costs in the context of industry-standard chip design tools..
- Our research shows that as the deployment scale expands, DAPT models based on smaller LLMs can produce substantial cost savings of more than 90-95% compared to larger SOTA LLMs. This translates into potential savings of millions of dollars for companies with extensive LLM usage aimed at increasing productivity in coding and software development tasks.
- Through our rigorous evaluation, we establish that domain-adaptive LLMs, despite their smaller size, can achieve comparable or even superior performance in coding assistance tasks compared to larger SOTA models. This highlights the importance of domain-specific fine-tuning and the efficiency of DAPT in capturing relevant knowledge and skills
- By shedding light on the trade-offs between performance and cost, we empower organizations to make strategic choices when integrating LLMs into their coding workflows. Our findings are relevant not only to the chip design industry, but also applicable to enterprises seeking to select appropriate LLMs for coding and software development assistance in various domains.

The paper is organized as follows. Section 2 outlines the domain adaptation and training methods used to include the DAPT, model alignment using SFT. Section 3 describes the experimental performance results for EDA tool code generation. Section 4 describes the detailed TCO analysis and comparison between DAPT-based LLM and SOTA larger LLMs. Finally, EDA coding examples used for benchmark evaluation are illustrated in the appendix.

## II. TRAINING METHODOLOGY FOR CHIPNEMO: ENHANCING ASIC PHYSICAL DESIGN WITH DOMAIN-ADAPTED LLM

LLM can be categorized into base models that are adaptable for various tasks, and specialized models refined through Supervised Fine-Tuning (SFT) or Reinforcement Learning from Human Feedback (RLHF) for task-specific performance. Training for LLMs involves data collection, self-supervised pre-training, and iterative fine-tuning, with the potential inclusion of reinforcement learning. Inference processes in LLMs consist of input processing, token prediction, selection strategies, and post-processing to generate coherent text. For example, GPT-3 and GPT-4 by OpenAI highlight the scalability of parameters and their impact on costs, with ChipNeMo, a domain-adapted model with 70 billion parameters, offering a cost-effective alternative for chip design applications.In the development of ChipNeMo, a specialized LLMs for ASIC physical design tasks, we base our approach on the LLaMA2 model, serving as the foundational architecture for subsequent domain-specific enhancements.

For the purpose of this study, we focused on a subset of the full dataset used for Domain Adaptive Pre-training (DAPT) and Supervised Fine-Tuning (SFT) of the ChipNeMo model. Although the complete data set encompasses a wide range of NVIDIA's internal documents, code snippets, and instructions, we specifically selected and utilized a portion of the data that is directly relevant to the Electronic Design Automation (EDA) tool code generation task. The token size of this curated subset was determined based on the documents and code samples pertaining to the specific EDA tool code generation task under investigation in this research work.

### A. Domain Adaptive Pre-training (DAPT)

To refine ChipNeMo's efficacy in domain-centric tasks, we employ Domain Adaptive Pretraining (DAPT). This technique extends the pre-training of the foundational LLM on a significant corpus of unlabeled data specific to the domain; in this instance, NVIDIA's internal chip design documentation and code. The objective of DAPT is to align the model's preexisting capabilities, derived from diverse and broad datasets, with the specialized vocabulary and context of the ASIC design domain. By immersing the model in domain-specific data, DAPT enhances the model's proficiency in processing and generating text that accurately reflects the linguistic patterns and technical terminology unique to chip design.

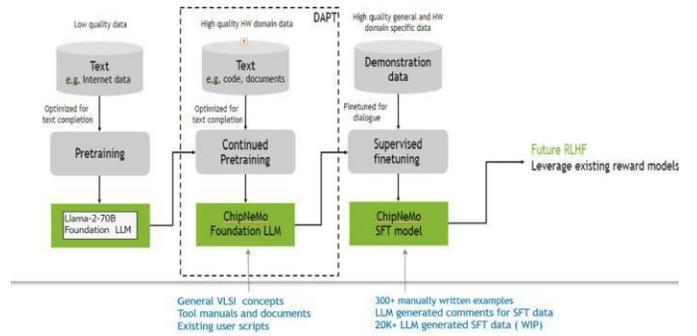

Fig. 1. *ChipNeMo-70B(LLM) SFT and DAPT Training flow*

For ChipNeMo's application to Electronic Design Automation (EDA) tool-related tasks in the chip design domain, the Domain Adaptive Pretraining (DAPT) phase required 100 GPU hours to process 120 million tokens. We used 24 billion tokens for DAPT; but only 120 million tokens are related to this task which included industry-standard Computer-Aided Design (CAD) tool documentation and relevant code snip- pets.This underscores the substantial computational resources necessary to adapt the language model to the specific domain.

### B. Supervised Fine-Tuning (SFT)

Following the Domain Adaptive Pre-training (DAPT) phase, ChipNeMo is subjected to Supervised Fine-Tuning (SFT), a process in which the model is further optimized using a more compact, task-oriented dataset containing labeled data. In particular, ChipNeMo underwent fine-tuning utilizing NVIDIA's proprietary sample Tool Command Language (Tcl) codes, which are associated with industry-standard Computer-Aided Design (CAD) tools. These tcl codes were meticulously annotated to support the SFT phase, allowing the model to learn from high-quality domain-specific examples and enhance its performance in chip design coding assistance tasks.

The SFT phase, which focused on training ChipNeMo for EDA tool-specific tasks, required 4 GPU hours to process 5 million tokens. This highlights the extensive computational resources and depth of training required to meticulously adapt the model for intricate chip design tasks, ensuring its proficiency in generating accurate and contextually relevant code snippets.

## III. PERFORMANCE COMPARISON OF LLMs IN ASIC DESIGN CODE GENERATION

. In the realm of ASIC design coding evaluation, the efficacy of general-purpose language models, such as those exemplified by ChatGPT, may not be fully aligned with the nuanced requirements of domain-specific tasks. This discrepancy has catalyzed the development and refinement of domain-adapted models such as ChipNeMo, which undergo fine-tuning on specialized datasets to encapsulate domain expertise, thus enhancing their precision, efficiency, and Total Cost of Ownership (TCO). This section delineates a comparative analysis of the ChipNeMo model against prominent large LLMs, namely ChatGPT-4 Turbo and Anthropic's Claude 3, focusing on their performance in chip design coding tasks.

### A. Models Under Comparison

- ChipNeMo: Tailored specifically for chip design, this domain-adapted LLM exemplifies the integration of domain-specific knowledge to augment coding evaluation tasks.
- Claude 3: Anthropic's latest iteration of a general-purpose LLM.
- ChatGPT-4 Turbo: An advanced conversational LLM developed by OpenAI.

### B. Task Description

To assess the performance of our model in generating Electronic Design Automation (EDA) scripts, we developed a benchmark consisting of coding tasks of medium to high complexity, as listed in appendix , inspired by real-world use cases encountered by engineers in their daily work. These tasks are significantly more challenging, necessitating multiple API calls. We enlisted human engineers to rate the accuracy of the generated scripts on a scale from 0 to 10. The benchmark currently comprises more than 25 Tool Command Language (tcl) based coding tasks for each complexity level (medium and hard). Ongoing efforts are focused on expanding the size and scope of these benchmarks to facilitate further assessment and refinement of the models.

### C. Description of benchmark

Performance evaluation benchmark encompassed several key metrics:

- **Accuracy in Coding Evaluation:** The models' proficiency was gauged through their ability to generate accurate tcl code snippets for Industry standard STA Tool from natural language prompts. A dataset comprising STA tcl code snippets (varying from moderate to high complexity) and related queries, annotated for common coding discrepancies, such as syntax and logical errors, served as the benchmark.
- **Hallucination Rate:** This metric quantified the models' propensity to produce outputs that either diverged from the input data or introduced factual inaccuracies.
- **Inference Speed:** The duration each model required to complete a standard set of coding evaluation tasks was meticulously recorded, providing insight into their operational efficiency.

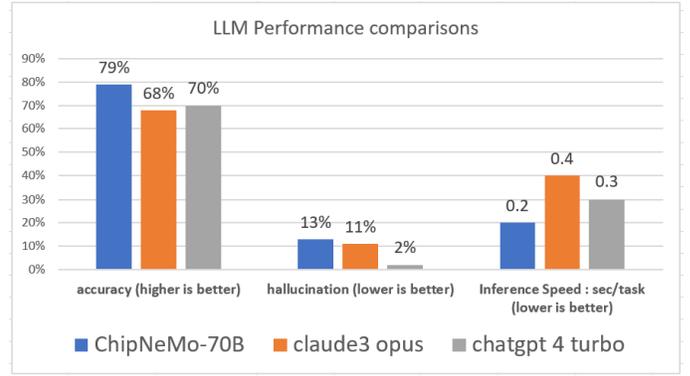

Fig. 2. *LLM Performance comparison*

### D. Comparative Performance Insights

- ChipNeMo demonstrated unparalleled performance in avioding syntax and logical errors, highlighting its proficiency in adhering to ASIC design coding standards and best practices. The model achieved an impressive accuracy rate of 79%, exceeding its counterparts, and exhibited the fastest inference speed among the models evaluated.
- Claude 3 was found to have the lowest accuracy at 68% and the high hallucination rate at 11%, indicating a possible misalignment with the specific demands of the chip design coding evaluation.
- ChatGPT-4 Turbo had a balanced profile with slightly lower accuracy (70%) compared to ChipNeMo but boasted the lowest hallucination rate at 2%, highlighting its robustness to generate consistent and accurate outputs.

This performance comparison elucidates the distinct advantages of employing a domain-adapted model such as ChipNeMo for ASIC design coding evaluations. ChipNeMo not only demonstrates superior accuracy in detecting coding errors but also maintains a competitive inference speed, affirming its tailored efficacy for chip design tasks. These findings align with the overarching goal of identifying economically viable and performance-efficient LLM solutions for specialized applications within the chip design industry.

## IV. TOTAL COST OF OWNERSHIP COMPARISON OF LLMs IN ASIC DESIGN CODING EVALUATION

### A. Cost Comparison Framework

To ascertain the economic viability of employing LLMs for ASIC design coding evaluation, we devised a comprehensive cost comparison methodology focusing on several pivotal aspects:

| Description | Lower workload | Average Workload | High Workload |
|---|---|---|---|
| Number of users | 175 | 175 | 175 |
| Queries/day | 15 | 30 | 50 |
| LLM usage life (days) | 120 | 120 | 120 |
| Total queries(M) | 0.3 | 0.6 | 1.1 |

TABLE I
*Workload calculation during a LLM life span (considering 6 month)*

| | |
|---|---|
| DAPT GPU(A100) hours | 100h |
| A100 cost/hour on Azure | $2 |
| total DAPT cost | $200 |
| SFT hours for CAD Tool | 4h |
| SFT cost | $8 |
| total DAPT+ SFT cost | $208 |

TABLE II
*DAPT and SFT training cost based on Azure cloud price*

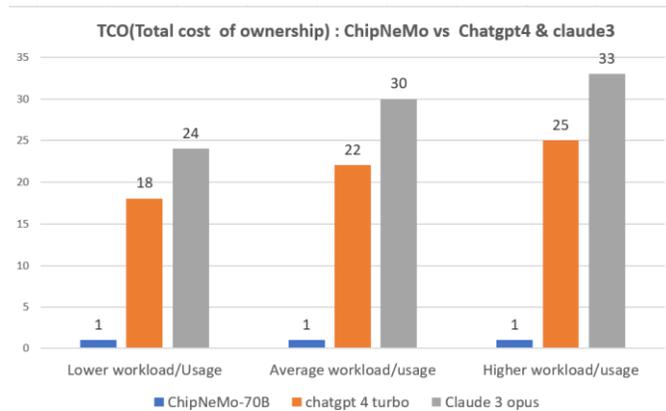

Fig. 3. *Comparative TCO Findings and Results for Domain-Adapted LLMs*

- **Model Parameter Size**: A comparative analysis of the trainable parameters in each model was conducted, recognizing its direct influence on the computational resources required for model inference.
- **Training Cost**: The expenditure associated with domain-adaptive pre-training (DAPT) and supervised fine-tuning (SFT) techniques was scrutinized, which is only applicable to ChipNeMo, a domain-adapted LLM designed for chip design tasks.
- **Inference and Operational Costs**: We projected the ongoing costs related to the deployment and maintenance of each model, including the inference cost incurred per API usage and query.

*B. Comparative TCO Findings and Results for Domain-Adapted LLMs in ASIC Design Coding Assistance:*

As depicted in Figure 3, Our analysis revealed that, on average, ChipNeMo-70B TCO is significantly lower—approximately 24-33 times less than Claude 3 and 18- 25 times less than ChatGPT-4 Turbo (based on workload).

We have revealed the remarkable potential of domain-adapted LLMs, such as ChipNeMo, to drastically reduce TCO by an impressive 90% to 95% compared to state-of-the-art SoTA LLMs such as Claude 3 Opus and ChatGPT-4 Turbo. The substantial cost reduction of domain-adapted LLMs is attributed to several key factors. First, the domain-specific training of ChipNeMo allows for a more compact model architecture, resulting in lower computational requirements and, consequently, reduced hardware and energy costs. Second, the specialized nature of ChipNeMo enables it to achieve higher performance levels with smaller training datasets, further minimizing data acquisition and annotation costs.

Furthermore, our analysis highlights a compelling trend: As the scale of deployment increases, the cost advantages of domain-adapted LLMs become even more pronounced. This is because the initial investment in domain-specific training and adaptation is amortized over a larger number of inference tasks, leading to a lower per-task cost. As a result, organi- zations with extensive chip design coding assistance need to benefit the most from adopting domain-adapted LLMs like ChipNeMo.

To put these findings in perspective, consider the case of a global manufacturer that achieved a reduction 95% in TCO by adopting managed services. Similarly, our research suggests that the adoption of domain-adapted LLMs in chip design coding assistance can lead to comparable, if not greater, cost savings. This underscores the immense potential of these specialized models to revolutionize the economics of LLM deployment in the chip design industry.

## V. CONCLUSION:

Our research has demonstrated the significant cost advantages of domain-adapted LLMs over SoTA alternatives in the context of chip design coding assistance. By achieving TCO reductions of 90% to 95%, models such as ChipNeMo offer a compelling value proposition for organizations seeking to optimize their LLM deployment costs while maintaining elevated levels of performance. As the scale of deployment grows, the economic benefits of domain-adapted LLMs become even more evident, making them an increasingly attractive option for the chip design industry.

| Items | lower Workload | Average workload | Higher workload |
|---|---|---|---|
| total EDA tool related Queries(M) | 0.3 | 0.6 | 1 |
| token/1 query | 1000 | 1000 | 1000 |
| total token(M) | 300 | 600 | 1000 |
| chatgpt inference cost $/1M | 30 | 30 | 30 |
| total chatgpt inference cost($) | 9000 | 18000 | 30000 |
| Claude inference cost ($/1M) | 40 | 40 | 40 |
| total claude 3 inference cost($) | 12000 | 24000 | 40000 |
| ChipNeMo inference cost ($/1M) | 1 | 1 | 1 |
| total ChipNeMo inference cost($) | 300 | 600 | 1000 |
| ChipNeMo training cost($) | 208 | 208 | 208 |
| total ChipNeMo cost($) | 508 | 808 | 1208 |
| TCO saving (ChipNeMo vs chatgpt 4) | 18 | 22 | 25 |
| TCO saving (ChipNeMo vs Claude 3) | 24 | 30 | 33 |

TABLE III
DETAILED TCO COMPARISON

## References

[1] L. Chen, Y. Chen, Z. Chu, W. Fang, T.-Y. Ho, Y. Huang, S. Khan, M. Li, X. Li, Y. Liang, Y. Lin, J. Liu, Y. Liu, G. Luo, Z. Shi, G. Sun, D. Tsaras, R. Wang, Z. Wang, X. Wei, Z. Xie, Q. Xu, C. Xue, E. F. Y. Young, B. Yu, M. Yuan, H. Zhang, Z. Zhang, Y. Zhao, H.-L. Zhen, Z. Zheng, B. Zhu, K. Zhu, and S. Zou, "The dawn of ai-native eda: Promises and challenges of large circuit models," 2024.

[2] B. Khailany, "Accelerating chip design with machine learning," in *Proceedings of the 2020 ACM/IEEE Workshop on Machine Learning for CAD*, ser. MLCAD '20. New York, NY, USA: Association for Computing Machinery, 2020, p. 33. [Online]. Available: https://doi.org/10.1145/3380446.3430690

[3] K. Chang, Y. Wang, H. Ren, M. Wang, S. Liang, Y. Han, H. Li, and X. Li, "Chipgpt: How far are we from natural language hardware design," 2023.

[4] M. Liu, N. Pinckney, B. Khailany, and H. Ren, "Invited paper: Verilogeval: Evaluating large language models for verilog code generation," in *2023 IEEE/ACM International Conference on Computer Aided Design (ICCAD)*, 2023, pp. 1–8.

[5] W. Fu, S. Li, Y. Zhao, H. Ma, R. Dutta, X. Zhang, K. Yang, Y. Jin, and X. Guo, "Hardware phi-1.5b: A large language model encodes hardware domain specific knowledge," 2024.

[6] J. Blocklove, S. Garg, R. Karri, and H. Pearce, "Chipchat: Challenges and opportunities in conversational hardware design," in *2023 ACM/IEEE 5th Workshop on Machine Learning for CAD (MLCAD)*, 2023, pp. 1–6.

[7] K. Chang, K. Wang, N. Yang, Y. Wang, D. Jin, W. Zhu, Z. Chen, C. Li, H. Yan, Y. Zhou, Z. Zhao, Y. Cheng, Y. Pan, Y. Liu, M. Wang, S. Liang, yinhe han, H. Li, and X. Li, "Data is all you need: Finetuning llms for chip design via an automated design-data augmentation framework," 2024.

[8] S. Thakur, B. Ahmad, Z. Fan, H. Pearce, B. Tan, R. Karri, B. Dolan-Gavitt, and S. Garg, "Benchmarking large language models for automated verilog rtl code generation," 2022.

[9] E. Nijkamp, B. Pang, H. Hayashi, L. Tu, H. Wang, Y. Zhou, S. Savarese, and C. Xiong, "Codegen: An open large language model for code with multi-turn program synthesis," 2023.

[10] M. Liu, N. Pinckney, B. Khailany, and H. Ren, "Verilogeval: Evaluating large language models for verilog code generation," 2023.

[11] H. Wu, Z. He, X. Zhang, X. Yao, S. Zheng, H. Zheng, and B. Yu, "Chateda: A large language model pow- ered autonomous agent for eda," *IEEE Transactions on Computer-Aided Design of Integrated Circuits and Systems*, pp. 1–1, 2024.

[12] M. Liu, T.-D. Ene, R. Kirby, C. Cheng, N. Pinckney, R. Liang, J. Alben, H. Anand, S. Banerjee, I. Bayraktaroglu, B. Bhaskaran, B. Catanzaro, A. Chaudhuri, S. Clay, B. Dally, L. Dang, P. Deshpande, S. Dhodhi, S. Halepete, E. Hill, J. Hu, S. Jain, A. Jindal, B. Khailany, G. Kokai, K. Kunal, X. Li, C. Lind, H. Liu, S. Oberman, S. Omar, G. Pasandi, S. Pratty, J. Raiman, A. Sarkar, Z. Shao, H. Sun, P. P. Suthar, V. Tej, W. Turner, K. Xu, and H. Ren, "Chipnemo: Domain-adapted llms for chip design," 2024.

APPENDIX

To maintain impartiality with all EDA vendors and comply with our internal policy, we present only the code/script queries used in this study without including the corresponding output code.

- Write Tcl proc to get All Unclocked Clock Pins and Their Drivers. If the clock pin has multiple sources, they are all listed on the same line
- Write Tcl proc to, finds failing paths and their respective startpoint and endpoint pairs. With this information, then reports available slack on the paths emanating from each failing endpoint and going into its associated startpoint
- Write Tcl proc to report all the cell which have negative delay. The output report prints the cell name, from-pin, to-pin, and also the rise and fall arc delays.
- Write Tcl code to get the nets which are connected to 3 or more load/reciever cells
- Write Tcl code to get the nets with multiple drivers cell
- Write Tcl code to get the unconstrained endpoints in the design
- write Tcl proc to get the interclock skew for a max timing path

Fig. 4. Sample EDA code generation queries